\documentclass[runningheads]{llncs}

\usepackage{graphicx}
\usepackage{booktabs}
\usepackage{capt-of}
\usepackage{wrapfig}
\usepackage{amsmath}
\usepackage{amssymb}

\begin{document}

\title{Sat2RealCity: Geometry-Aware and Appearance-Controllable 3D Urban Generation from Satellite Imagery}

\titlerunning{Sat2RealCity: Satellite-to-3D Urban Generation}

\author{
Xinliang Wang\textsuperscript{*} \and
Yijie Kang\textsuperscript{*} \and
Zhenyu Wu \and
Yifeng Shi\textsuperscript{\ensuremath{\dagger}}
}

\authorrunning{X. Wang et al.}

\institute{
KE Holdings Inc., Beijing, China\\
\email{\{wangxinliang008, kangyijie001, wuzhenyu018\}@ke.com}\\
\email{shiyifeng@tju.edu.cn}
}

\maketitle

\begingroup
\renewcommand{\thefootnote}{}
\footnotetext{
\textsuperscript{*}Equal contribution.
\hspace{0.5em}
\textsuperscript{\ensuremath{\dagger}}Corresponding author.
}
\addtocounter{footnote}{-1}
\endgroup

\begin{abstract}
3D urban generation from satellite imagery is an important task for scalable digital twins and real-world simulation environments. Existing approaches primarily rely on scene-level generation paradigms, which often require large-scale 3D city assets and struggle with controllability, geographic alignment, and realistic appearance grounding in real-world urban environments. To address these limitations, we present \textbf{Sat2RealCity}, a grounded urban generation framework that leverages object-level 3D generative priors for scalable city synthesis from satellite imagery. Our framework decomposes cities into geographically grounded building entities, enabling the reuse of pretrained object-level 3D generative priors while preserving real-world spatial structures. Supported by our constructed \textbf{BuildVerse3D} dataset, (1) we introduce an OpenStreetMap (OSM)-guided spatial grounding strategy to inject geospatial constraints into the 3D generation process; (2) we design an appearance-guided controllable generation mechanism for realistic architectural appearance and regional style consistency; and (3) we construct an MLLM-powered semantic pipeline for regional appearance understanding and semantic-aware appearance synthesis. Extensive experiments demonstrate that Sat2RealCity achieves strong geographic alignment, regional stylistic consistency, and plausible urban asset synthesis compared with existing urban generation and 3D asset generation approaches.
\keywords{3D Urban Generation \and Satellite Imagery \and Building-Entity 3D Generation}
\end{abstract}

\section{Introduction}
\label{sec:intro}
Generating large-scale 3D urban environments from satellite imagery is valuable for digital twins, urban visualization, and large-scale spatial content creation. Advances in dense visual prediction, object detection, monocular 3D perception, and transportation foundation models have improved the perception and understanding of complex urban environments
\cite{xia2024vit,wang2025rt,jinrang2023monouni,li2024monolss,shi2023open,chen2023transiff,kong2023dusa,jia2024ropebev}.
Complementary to these perception-centric efforts, recent progress in 3D spatial understanding, reconstruction, and generative world modeling has advanced the modeling of coherent 3D environments
\cite{xing2025doremi,xu2025cruise,xing2026adaptsplat,wang2026artifactworld,jia2026panoworld,li2026pano2world,jia2026you}.
Sat2RealCity focuses on a distinct but related problem: generating explicit, editable, and geographically grounded urban assets from overhead imagery. Nevertheless, existing urban generation approaches still face several practical limitations. One line of work~\cite{hua2025sat2city,xie2024citydreamer,liu2025earthcrafter} focuses on scene-level 3D generation using dense representations such as NeRF~\cite{mildenhall2021nerf}, 3DGS~\cite{kerbl20233d}, or heavy voxel grids. While these methods demonstrate promising large-scale urban synthesis capabilities, they often require large-scale 3D city assets and struggle with controllability, geographic alignment, and realistic appearance grounding in real-world urban environments. Another line of work~\cite{lu2020geometry,shi2022geometry,li2021sat2vid,qian2023sat2density,xu2024geospecific} focuses on cross-view image synthesis from satellite imagery. While achieving impressive 2D view synthesis quality, these methods do not model editable 3D geometry or support free-view rendering.

Meanwhile, object-level 3D generation has rapidly advanced in recent years, demonstrating strong generative priors and high-quality 3D asset synthesis capabilities. However, effectively extending these mature object-centric priors toward grounded urban generation remains largely unexplored. To address this gap, we propose \textbf{Sat2RealCity}, a building-entity-level framework for grounded 3D urban generation. By decomposing urban environments into individual architectural entities, our framework leverages reusable object-level 3D generative priors. Compared with holistic scene-level generation, this decoupled approach enables compositional urban synthesis while preserving real-world geographic structures, providing a practical foundation for scalable real-world urban creation.

\begin{figure}[t]
\centering
\includegraphics[width=0.7\linewidth]{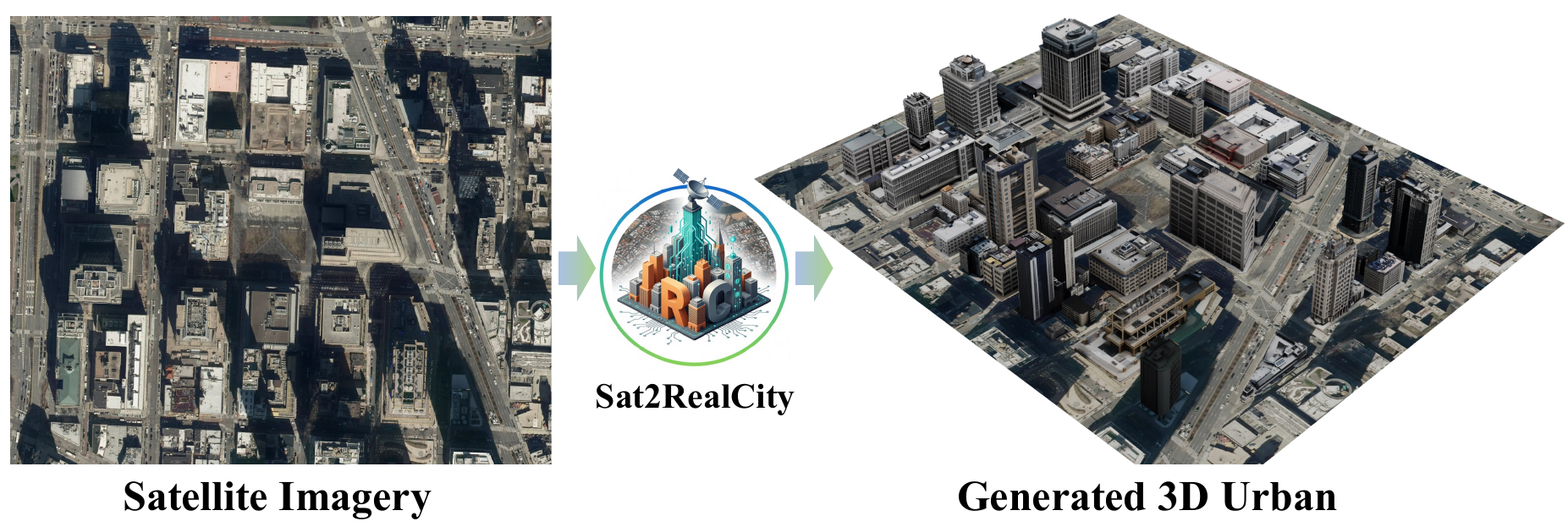}
\vspace{-3pt}
\caption{We present \textbf{Sat2RealCity}, a framework for generating explicit 3D urban assets with controllable geometry and appearance from real-world satellite imagery.}
\label{fig:first}
\end{figure}

To support this framework, we construct \textbf{BuildVerse3D}, a high-quality dataset addressing the scarcity of 3D architectural assets. During inference, Sat2RealCity improves geographic grounding and regional stylistic consistency through three components: (1) an OpenStreetMap (OSM)-based spatial priors strategy that aligns generation with real-world geospatial footprints; (2) an appearance-guided controllable mechanism that maintains regional stylistic coherence while preserving building-specific geometry; and (3) an MLLM-powered semantic pipeline that summarizes shared regional semantic attributes from visually similar buildings and converts them into consistent appearance references for scalable urban generation.

Our main contributions are summarized as follows:

\begin{itemize}

\item We propose a building-entity-level framework for grounded 3D urban generation that leverages reusable object-level generative priors for compositional urban synthesis. Our framework enables scalable urban generation while preserving geographic consistency and building-level structural flexibility.

\item We design an automated generation pipeline integrating an OSM-based spatial grounding strategy, an appearance-guided controllable mechanism, and an MLLM-powered semantic module. Together, they unify geographic grounding with regional stylistic consistency for real-world urban generation.

\item We construct BuildVerse3D, a high-quality synthetic 3D building dataset that provides rich object-level generative priors for scalable urban generation.

\end{itemize}
\section{Related Work}
\label{sec:related}
\begin{figure}[t] 
    \centering
    \includegraphics[width=\textwidth]{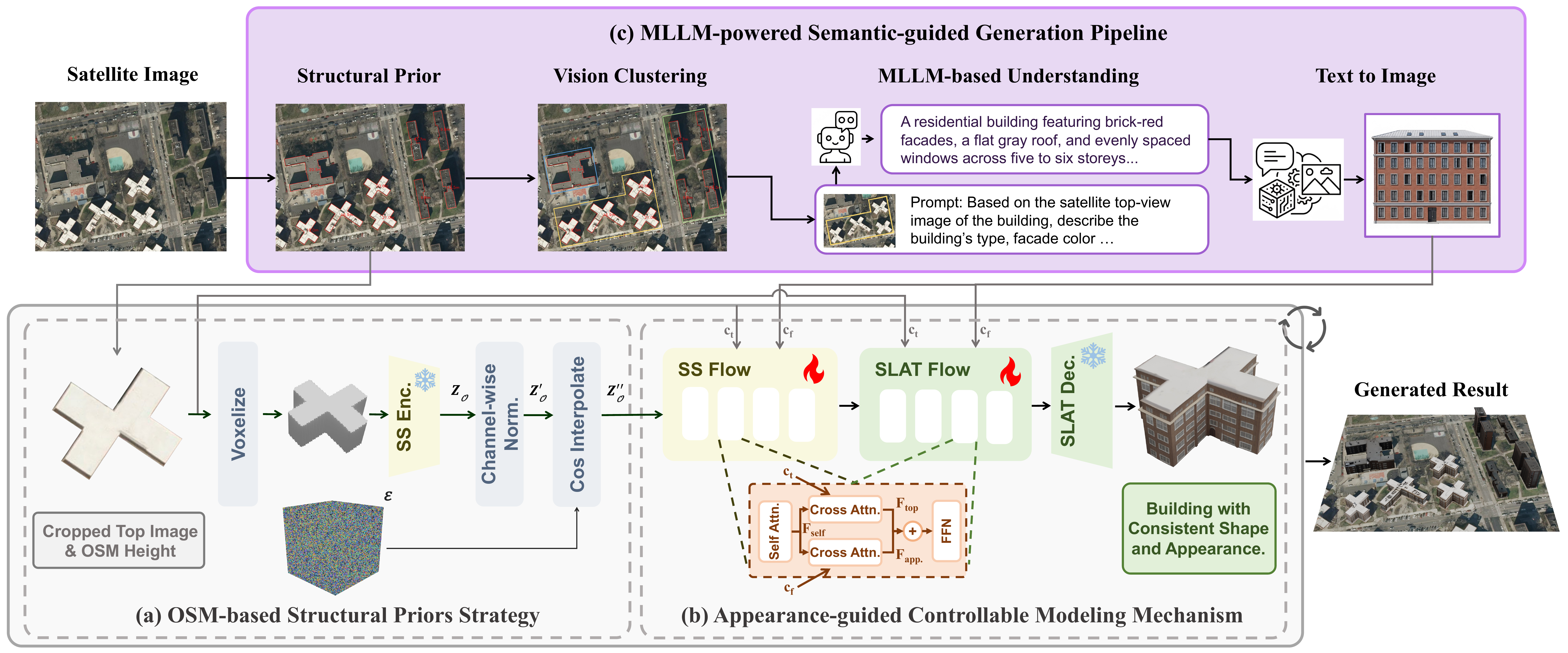} 
    \caption{\textbf{The overview of Sat2RealCity.} 
    (a) The \textbf{OSM-based Structural Priors Strategy} converts OSM data into a fused geometric prior $Z''_{\mathcal{O}}$.
    (b) The \textbf{Appearance-guided Modeling Mechanism} uses $Z''_{\mathcal{O}}$, the top view feature $c_t$, and a frontal appearance image feature $c_f$ to generate the 3D building.
    (c) The \textbf{MLLM-powered Generation Pipeline} provides geometric priors and the frontal-view image for modules (a) and (b).
    Finally, all generated buildings are assembled into the urban scene using their original OSM coordinates.
    }
    \label{fig:framework}
\end{figure}
\subsection{3D Urban Generation and Cross-View Synthesis}

3D urban generation aims to synthesize city-scale scenes with realistic geometry and appearance, supporting urban planning and simulation. Procedural techniques~\cite{deng2024citycraft} generate structured layouts but lack fine-grained realism. Recent data-driven approaches leverage semantic maps~\cite{hua2025sat2city,xie2024citydreamer} or multi-modal priors~\cite{liu2025earthcrafter} (including concurrent multi-view diffusion~\cite{yao2025magiccity}) to produce realistic urban layouts. However, even when attempting compositional generation to decouple scene elements, these methods predominantly rely on dense radiance fields or massive point clouds. This inherently limits their ability to produce explicit, individually editable 3D architectural assets. Crucially, when utilizing satellite imagery, related research focuses on cross-view image synthesis~\cite{qian2023sat2density,xu2024geospecific} to translate top-down inputs into street-level views. While achieving impressive 2D visual quality, they are view-synthesis frameworks rather than 3D urban scene generators. Their implicit or 2D-based representations cannot produce explicit geometric meshes, preventing deployment in downstream rendering engines. Furthermore, they are optimized for ground-level camera trajectories, failing to support unconstrained free-viewpoint rendering. In contrast to cross-view synthesis methods, Sat2RealCity tackles explicit 3D urban asset generation. We adopt a building-entity-based paradigm to reconstruct individual, editable 3D meshes that strictly align with real-world satellite footprints.

\subsection{3D Object Generation}

Early 3D object generation relied on optimization techniques like Score Distillation Sampling (SDS)~\cite{poole2022dreamfusion}, which suffers from slow speeds and inconsistent geometry. To achieve efficient feed-forward generation~\cite{yang2024hunyuan3d,tang2024lgm,xu2024instantmesh,zhang2024gs}, the community explored two dominant paths for 3D latent representations. The first relies on vector-set (VecSet) representations~\cite{li2025triposg,hunyuan3d22025tencent,hunyuan3d2025hunyuan3d,lai2025hunyuan3d}, originating from 3DShape2VecSet~\cite{zhang20233dshape2vecset}. This approach encodes a 3D shape as an unordered set of vectors, which naturally suits Transformer architectures. Another research line focuses on the Voxel-VAE paradigm, including XCube~\cite{ren2024xcube}, TRELLIS~\cite{xiang2025structured} and SparseFlex~\cite{he2025sparseflex}. While these methods generate individual objects, producing complete, real-world-aligned urban scenes from satellite imagery remains challenging. Sat2RealCity adapts these advanced latent representations into a building-entity-level paradigm. By integrating OSM spatial priors and appearance-guided control, our framework explicitly generates editable polygon meshes, bridging the gap between single-object generation and scalable urban simulation.
\section{Method}
\label{sec:method}
\subsection{Preliminary and Overview}
\label{sec:prelim}

Our framework builds upon TRELLIS~\cite{xiang2025structured}, which generates explicit 3D assets via a two-stage Rectified Flow Transformer. It first synthesizes a sparse voxel structure $\{p_i\}^L_{i=1}$ (SS Flow), followed by a structured latent $z=\{(z_i,p_i)\}^L_{i=1}$ (SLAT Flow) that can be decoded into explicit meshes. Both stages accept DINOv2~\cite{oquab2023dinov2} features as conditions and are optimized by minimizing the Conditional Flow Matching (CFM) objective: $\mathcal{L}_{CFM}(\theta)=\mathbb{E}_{t,\mathbf{x_0},\epsilon}\|\mathbf{v_\theta}(\mathbf{x},t)-(\epsilon-\mathbf{x_0})\|^2_2$.

To bypass the prohibitive rendering costs and geometric entanglement of monolithic implicit fields, we extend this generative paradigm for building-entity-level explicit 3D generation (Fig.~\ref{fig:framework}). Specifically, our Sat2RealCity model is trained on the proposed BuildVerse3D dataset (Sec.~\ref{sec:dataset}). During inference, it systematically injects OSM-guided structural priors (Sec.~\ref{sec:osm_priors}) and MLLM-generated appearance conditions (Sec.~\ref{sec:appearance_control}) into the flow matching process, autonomously bridging individual explicit architectural meshes into a geometry-aware, stylistically consistent, and real-time renderable 3D urban scene.

\subsection{BuildVerse3D Dataset}
\label{sec:dataset}

While 3D datasets have surged, none provide dedicated, high-fidelity 3D building assets suitable for advanced generation. Existing collections suffer from limitations: (1) \textbf{General 3D repositories}~\cite{objaverseXL,collins2022abo,fu20213d} lack architectural focus and topological consistency; (2) \textbf{GIS and urban-scale datasets}~\cite{seto2023role,wenrealcity3d} prioritize macroscopic coverage, yielding low-polygon generic structures with baked-in lighting; and (3) \textbf{Handcrafted architectural datasets}~\cite{selvaraju2021buildingnet,lin2022capturing} remain limited in scale and stylistic diversity due to prohibitive manual modeling costs.

\begin{wrapfigure}{r}{0.42\textwidth}
    \vspace{-10pt}
    \centering
    \includegraphics[width=\linewidth]{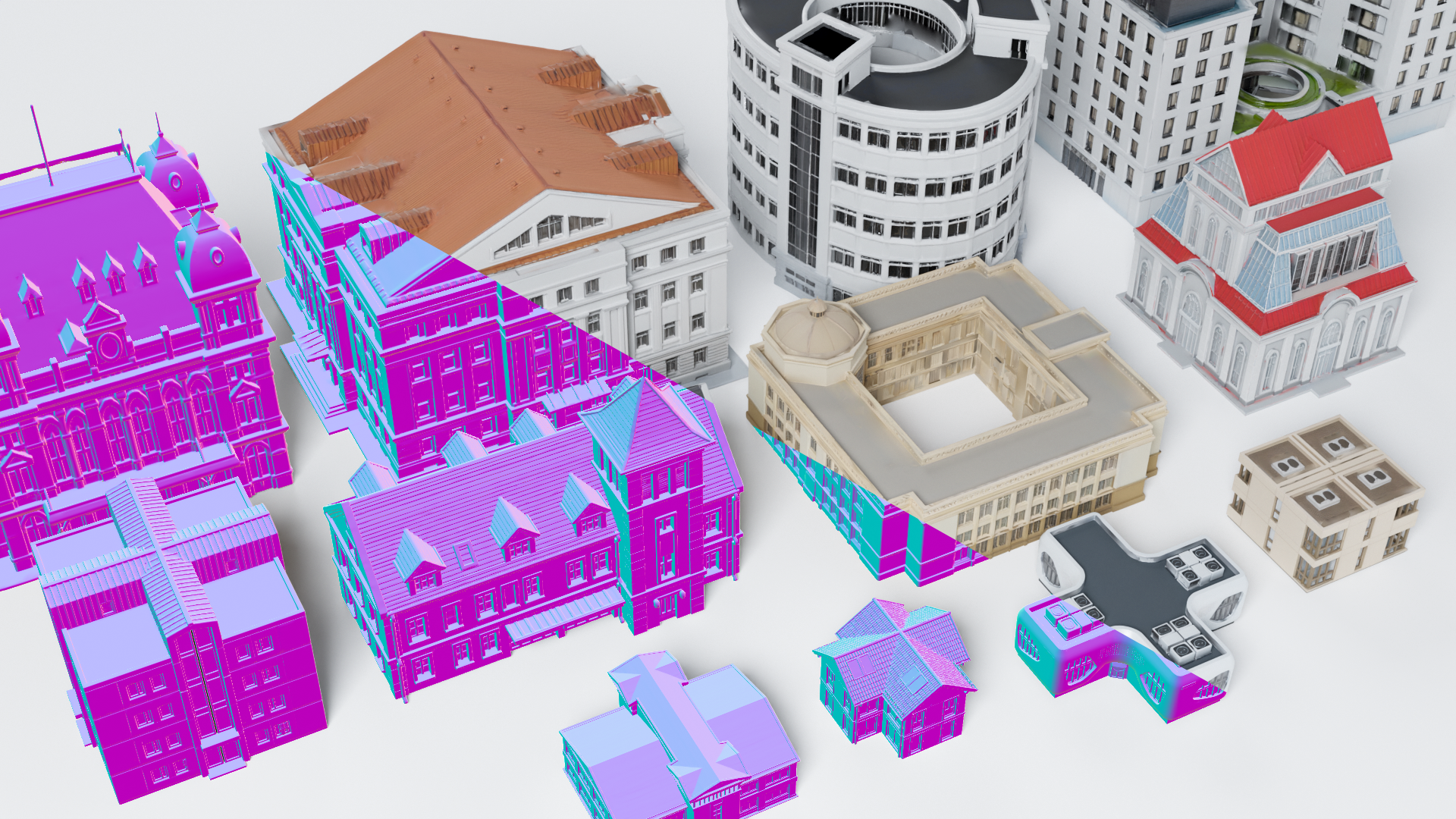}
    \caption{\textbf{Examples of the BuildVerse3D Dataset}, featuring high-fidelity, stylistically diverse, and cleanly meshed architectural assets.}
    \label{fig:dataset_fig}
    \vspace{-10pt}
\end{wrapfigure}
To bridge this gap and empower our entity-level generation paradigm, we construct \textbf{BuildVerse3D}, a novel open-source dataset comprising 11,579 high-quality synthetic building models (Fig.~\ref{fig:dataset_fig}). Overcoming prior limitations, our dataset guarantees: \textit{(i) Artist-level geometry} featuring intricate structural details and clean meshes; \textit{(ii) Shadowless pure albedo textures} readily supporting relighting tasks; and \textit{(iii) Balanced diversity} across a broad spectrum of architectural categories and styles.

To ensure semantic coherence and visual fidelity, BuildVerse3D is constructed via an automated text-to-image-to-3D pipeline. We first design a hierarchical prompt generator validated by LLMs, utilize powerful text-to-image models (Imagen~4) to extract 2D visual priors, and finally employ a 3D generation model (Hunyuan3D~2.5~\cite{lai2025hunyuan3d}) to obtain the assets. The detailed construction protocol is provided in the supplementary material.

\subsection{OSM-based Structural Priors for Geometry and Pose Alignment}
\label{sec:osm_priors}

Generating 3D buildings from satellite imagery lacks height, fine geometry, and pose cues. To resolve this inherent ambiguity, we propose the \textbf{OSM-based structural priors strategy} (Fig.~\ref{fig:framework}a) to provide explicit geometric guidance.

Crucially, we adopt distinct structural proxies during training and inference. During training, we use downsampled versions of 3D asset voxels as proxies; during inference, we extrude real OSM building footprints into coarse volumetric proxies. These proxies are then encoded via a pretrained Sparse Structure (SS) VAE encoder~\cite{xiang2025structured} into a prior latent $Z_{\mathcal{O}}$. Following a channel-wise normalization to align the distribution with a Gaussian prior ($Z'_{\mathcal{O}}$), we perform Cosine Geometric Interpolation with Gaussian noise $\epsilon \sim \mathcal{N}(0, I)$ to balance deterministic cues with generative stochasticity:
\begin{equation}
Z''_{\mathcal{O}} = \cos\left(\frac{\lambda\pi}{2}\right)Z'_{\mathcal{O}} + \sin\left(\frac{\lambda\pi}{2}\right)\epsilon,
\label{eq:cosine_interp}
\end{equation}
where $\lambda$ controls the strength of the structural prior. The fused latent $Z''_{\mathcal{O}}$ then serves as the initialization for the SS Flow Transformer. Furthermore, during training, we stochastically sample multiple Levels of Detail (e.g., LOD 0 bounding boxes and LOD 1 extruded footprints) paired with varying noise intensities to prevent overfitting and enhance generalization. In contrast to Sat-Skylines~\cite{jin2025sat}, which employs structural priors mainly for coarse shape and height conditioning, our method leverages an implicit but crucial advantage: joint geometry-aware and geospatial alignment. Because OSM footprints are geospatially registered with satellite imagery, the encoded latent inherently captures both absolute orientation, building contour, aspect ratio, and relative elevation. Consequently, Sat2RealCity directly generates 3D structures that are structurally faithful and automatically pose-aligned to real-world coordinates. 

\subsection{Appearance-Guided Controllable Regional Style Consistency}
\label{sec:appearance_control}

While OSM priors enforce spatial alignment, generating buildings solely from individual top-view crops yields visually fragmented neighborhoods with excessive appearance variance. Conversely, relying exclusively on a shared frontal view destroys their unique satellite-derived footprints and roof structures. Real-world urban blocks typically share architectural styles (e.g., facade materials, colors) despite varying geometries. To resolve this dilemma, we introduce an \textbf{appearance-guided controllable modeling mechanism} (Fig.~\ref{fig:framework}b). By conditioning on both perspectives, it enforces a unified regional style via a shared frontal appearance condition while strictly preserving individual geometric layouts via top-view crops.

We propose parallel cross-attention pathways, which are incorporated within each SS Flow and SLAT Flow transformer block, following the self-attention and preceding the feed-forward network (FFN), as illustrated in Fig.~\ref{fig:framework}:

\noindent \textbf{Structure Pathway:} conditioned on the top (satellite) image, which is rendered from a top view of the 3D asset during training and cropped from real satellite imagery based on OSM footprints during inference, enforcing geometric fidelity and spatial alignment. Formulated as:
\begin{equation}
F_{top}=CrossAttn(Q(F_{self}),K(c_{t}),V(c_{t}));
\label{eq:top_cross}
\end{equation}
\noindent \textbf{Appearance Pathway:} conditioned on a frontal image, which is rendered from a frontal view of the 3D asset during training and generated by our MLLM-based pipeline (Sec.~\ref{sec:mllm}) during inference, guiding stylistic coherence across buildings. Formulated as:
\begin{equation}
F_{appearance}=CrossAttn(Q(F_{self}),K(c_{f}),V(c_{f})),
\label{eq:front_cross}
\end{equation}
where $F_{self}$ is the output of the self-attention layer, $c_t$ and $c_f$ are respectively the top view and the frontal-view DINOv2~\cite{oquab2023dinov2} feature, $Q(\cdot)$,$K(\cdot)$ and $V(\cdot)$ are linear layers respectively for query, key and value projection. To preserve the pretrained capacity of the foundation model, the weights of both cross-attention pathways are initialized from the pretrained weights. 

The outputs of both pathways, \( F_{\text{top}} \) and \( F_{\text{appearance}} \), are fused by simple element-wise averaging before being passed to the FFN.

This dual-pathway design elegantly disentangles geometry and style. Since the transformer operates on Structured Latent (SLAT) representations \cite{xiang2025structured} \allowbreak equipped with explicit 3D positional information $p_i$, the model learns a spatially-dependent routing policy. It prioritizes the top-view signal $F_{top}$ for spatial indices corresponding to roofs, while prioritizing the frontal-view signal $F_{appearance}$ for facades. Consequently, when multiple buildings share the same $c_f$ but possess unique $c_t$, the mechanism seamlessly applies a unified architectural style across the neighborhood without violating footprints, as validated in Sec.~\ref{subsec:exp_consistency}.

\subsection{MLLM-powered Semantic-guided Generation Pipeline}
\label{sec:mllm}

We introduce an \textbf{MLLM-powered semantic-guided generation pipeline} that bridges satellite imagery with structure- and appearance-controllable urban generation through multimodal understanding. As illustrated in Fig.~\ref{fig:framework}(c), after retrieving building polygons and heights from OSM, we perform a \textit{vision-based clustering} step to capture visual similarities. Specifically, we employ DINOv3~\cite{simeoni2025dinov3} to extract top-view features for each building, yielding a feature vector $\mathbf{v}_i \in \mathbb{R}^{1024}$. This is concatenated with the building height $h_i$ to form a joint representation $\mathbf{f}_i = [\mathbf{v}_i; h_i] \in \mathbb{R}^{1025}$. These representations are then clustered using HDBSCAN to group buildings sharing similar geometric and visual characteristics. For each cluster, we conduct an \textit{MLLM-based semantic understanding} stage. The top-view images of all buildings within a cluster are jointly input to the multimodal large language model. Guided by a structured prompt querying attributes like building type, exterior color, material, and window layout, the MLLM produces a concise textual description summarizing the shared visual semantics. Finally, the generated descriptions are passed to a \textit{text-to-image} (T2I) model to synthesize corresponding appearance reference images. These generated appearance maps serve as controllable style guidance for the downstream 3D urban generation framework.

\begin{table}[t]
\centering
\footnotesize
\caption{Quantitative comparison of 3D urban generation. `T' denotes satellite top-view input; `F' indicates front-view input from the MLLM understanding pipeline.}
\setlength{\tabcolsep}{3pt} 
\begin{tabular}{lccccc} 
\toprule
Method & Input & CD$\downarrow$ & F$\uparrow$ & CLIP$\uparrow$ & DINOv2 sim$\uparrow$ \\
\midrule
        TRELLIS & T & 0.2803 & 0.1835 & 0.7862 & 0.6587 \\
        Step1x-3D & T & 0.2364 & 0.2073 & 0.7505 & 0.4522 \\
        Hunyuan3D-2.1  & T & 0.1402 & 0.3650 & 0.7626 & 0.4413 \\
        TRELLIS-MV-M & T+F & 0.0625 & 0.4699 & 0.7989 & 0.7899 \\
        TRELLIS-MV-S & T+F & 0.0879 & 0.4727 & 0.8120 & 0.7756 \\
        Sat2RealCity~(Ours) & T+F & \textbf{0.0138} & \textbf{0.8890} & \textbf{0.8563} & \textbf{0.8114} \\
\bottomrule
\end{tabular}
\label{tab:region_quantitative}
\end{table}

\begin{figure}[t] 
    \centering
    \includegraphics[width=\textwidth]{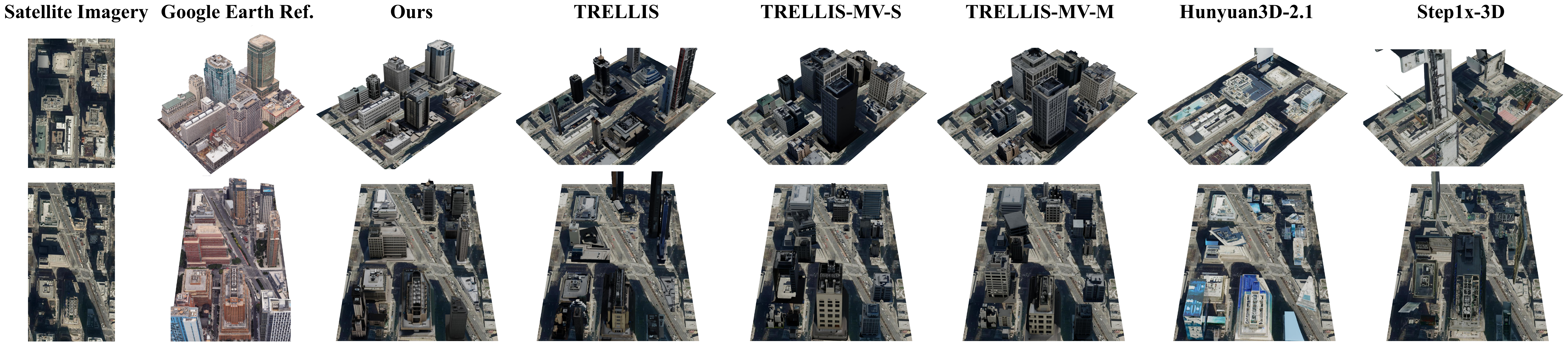} 
    \caption{Qualitative visualization of 3D urban generation.} 
    \label{fig:region}
\end{figure}

\section{Experiments}
\label{sec:exp}
\subsection{Evaluation Protocol Overview.}
Unlike existing city generation methods that synthesize dense 3D representations or 2D renderings, Sat2RealCity uses a building-entity-level 3D generation paradigm. Our core innovation lies in adapting state-of-the-art 3D object foundation models to generate explicit, real-world-aligned urban assets. Therefore, to rigorously demonstrate our superiority in precise geometric control and regional style consistency, our primary evaluation (Sec.~\ref{subsec:urban_gen}-Sec.~\ref{subsec:exp_consistency}) explicitly compares against object-centric baselines (e.g., TRELLIS, Hunyuan3D). For completeness, macroscopic comparisons with city scene generators are presented in Sec.~\ref{subsec:exp_citygen_compare}.

\subsection{Implementation details}
Our building-entity-based generation model was initialized with TRELLIS-image-large weights and fine-tuned on our 3D building dataset using a learning rate of 0.0001, a batch size of 32, and 40K training steps on 8 NVIDIA A100 GPUs. During training, the OSM-based structural priors module randomly adopted LOD 0 or LOD 1. During inference, satellite imagery and OSM height data were collected from Mapbox and the Overture Maps Foundation, respectively. The clustering module utilized HDBSCAN. The MLLM generating textual appearance descriptions was implemented using Qwen3-VL~\cite{Qwen2.5-VL}, while the model synthesizing building appearances employed FLUX-dev~\cite{labs2025flux}. Detailed prompts are provided in the supplementary material. During inference, the hyperparameter $\lambda$ in the OSM-based structural priors module was fixed to 0.5, and exclusively LOD 1 geometric priors were utilized.

\subsection{Evaluation Data and Metrics}
\label{subsec:data_metrics}
\noindent \textbf{Data.} We evaluate our framework on 111 real-world urban regions with high-quality satellite coverage, totaling 10~$km^2$. The evaluation covers geometry and appearance.
For geometry-related evaluation, we use OSM building footprints with height information as reference structural annotations, which characterize coarse building envelopes rather than detailed architectural geometry. Google Earth Studio city mesh renderings are used as appearance references.

\noindent \textbf{Metrics.}
To evaluate \emph{geometric consistency}, we compare point clouds from the generated regions against the OSM-derived reference structures using the Chamfer Distance (CD) and F-score. 
CD measures average bidirectional distance, and F-score reflects points within a distance threshold.

For \emph{appearance realism}, we adopt the CLIP Score~\cite{radford2021learning}, defined as the cosine similarity between CLIP (ViT-L/14) embeddings of our rendered city views and those of the Google Earth Studio reference. 
This metric quantifies the perceptual alignment between generated textures and realistic visual appearances.
Additionally, we employ the DINOv2 sim, calculated as the cosine similarity between DINOv2 (dinov2-vitl14-reg) ~\cite{oquab2023dinov2} features of the generated and reference views. This metric evaluates fine-grained structural and textural fidelity.

\subsection{Entity-Assembled Urban Generation}
\label{subsec:urban_gen}
\noindent \textbf{Settings.} 
Following our entity-level evaluation protocol, we evaluate several advanced 3D object generation models, including TRELLIS~\cite{xiang2025structured} (single-image version TRELLIS and multi-view variant TRELLIS-MV). The multi-view variant includes two strategies: multi-diffusion-based (TRELLIS-MV-M) and stochastic sampling (TRELLIS-MV-S). We also evaluate Hunyuan3D-2.1~\cite{hunyuan3d2025hunyuan3d} and Step1X-3D~\cite{li2025step1x} under the same building-entity-based setting. We evaluate all methods on both geometric consistency and appearance realism at the regional level, using the data and metrics introduced in Sec.~\ref{subsec:data_metrics}. The input condition images for TRELLIS-MV include both building rooftop views and appearance guidance images, while other baseline methods use only building rooftop views as input.

\begin{figure}[t]
    \centering
    \begin{minipage}[t]{0.50\textwidth}
        \vspace{0pt} 
        \centering
        \captionof{table}{Quantitative comparison of geometric fidelity. `T' denotes the satellite top-view input, while `F' denotes the front-view input rendered from 3D assets.}
        \label{tab:geo_quantitative}
        
        \footnotesize 
        \setlength{\tabcolsep}{4pt} 
        
        \begin{tabular}{lccc}
        \toprule
        Method & Input & CD$\downarrow$ & F$\uparrow$  \\
        \midrule
        TRELLIS & T & 0.0508 & 0.5778  \\
        Step1x-3D & T & 0.0230  & 0.6921 \\
        Hunyuan3D-2.1  & T &  0.0202 & 0.7099 \\
        Ours-T & T & \textbf{0.0150} & \textbf{0.7871} \\
        \midrule
        TRELLIS-MV-M & T+F & 0.0184 & 0.7929  \\
        TRELLIS-MV-S & T+F & 0.0199 & 0.7763  \\
        Ours & T+F & \textbf{0.0118} & \textbf{0.8554} \\
        \bottomrule
        \end{tabular}
    \end{minipage}
    \hfill
    \begin{minipage}[t]{0.44\textwidth}
        \vspace{0pt} 
        \centering
        \includegraphics[width=\linewidth]{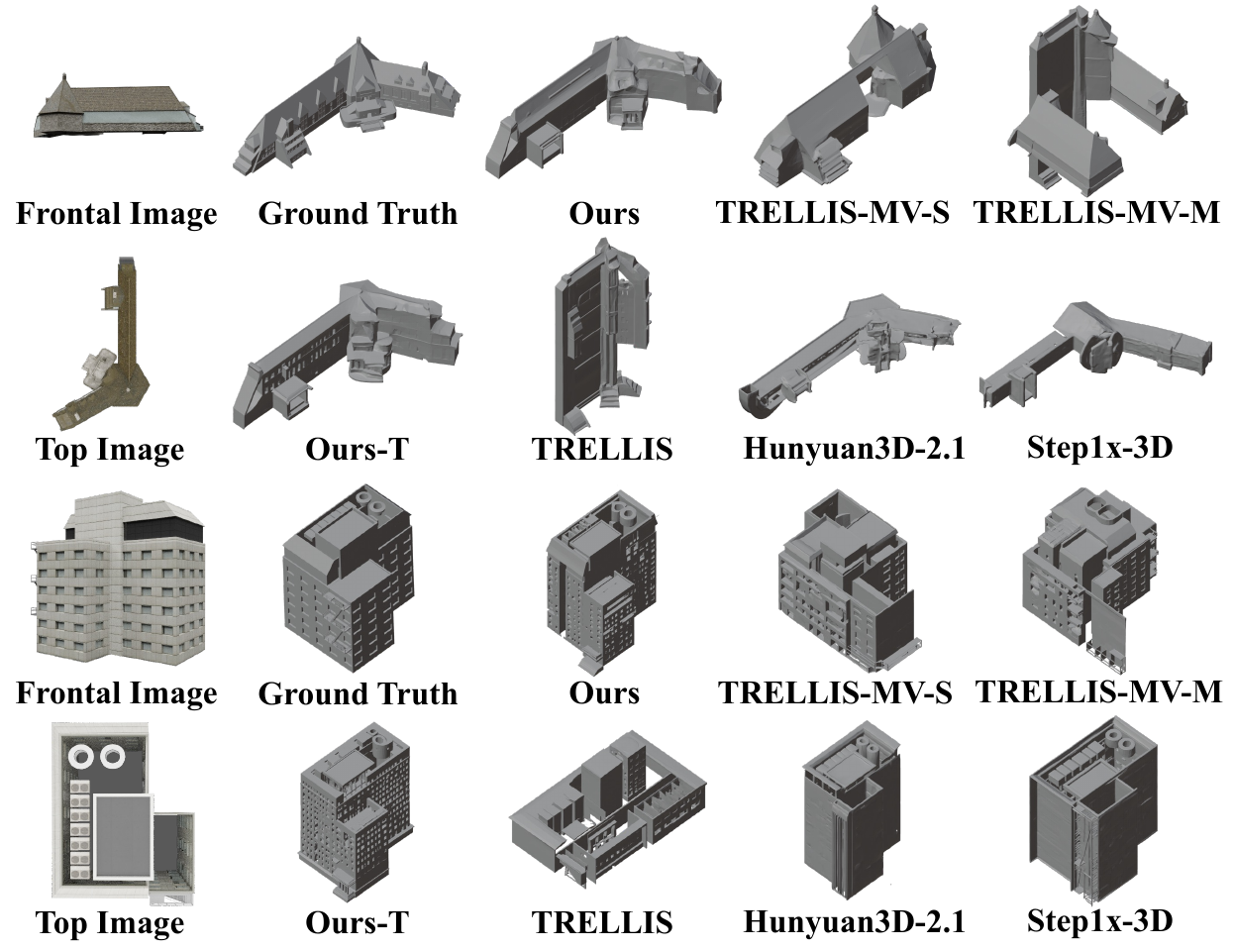}
        \captionof{figure}{Qualitative visualization of geometric fidelity.}
        \label{fig:geo}
    \end{minipage}
\end{figure}

\noindent \textbf{Quantitative Results.}
Tab.~\ref{tab:region_quantitative} summarizes the results. For \emph{geometric consistency}, our method achieves the lowest CD and highest F-score among baselines, demonstrating superior spatial fidelity. 
This stems from our OSM-based structural priors explicitly encoding real-world constraints. Baselines lacking pose control tend to produce misaligned or rotated structures. 
Even after post-hoc rotational registration of all baselines, Sat2RealCity, without correction, achieves higher geometric accuracy, validating the effectiveness of our implicit rotational pose alignment integrated directly into generation.

For \emph{appearance realism}, our model achieves the highest CLIP Score and DINOv2 sim, indicating closest alignment with real-world visuals. This results from our dual-pathway cross-attention, which preserves structural accuracy from satellite priors while incorporating plausible architectural styles from MLLM-generated appearance conditions. Baselines fail to balance structural control with photorealistic facades, yielding lower visual fidelity.

\noindent \textbf{Qualitative Results.}
As illustrated in Fig.~\ref{fig:region}, our generated city regions exhibit both precise geospatial alignment and coherent regional style. 
Buildings are correctly oriented with respect to their OSM footprints and maintain stylistic harmony within each neighborhood. 
In comparison, baseline results often show inconsistent orientations, distorted roof geometry, or abrupt texture transitions across adjacent buildings. 
These qualitative observations further corroborate our quantitative findings, demonstrating that Sat2RealCity effectively achieves geometry-faithful and style-consistent regional urban synthesis.

\subsection{Ablation Study on Geometric Fidelity}
\label{sec:exp_geometry_ablation}
\noindent \textbf{Settings.}  
To further assess the contribution of our OSM-based structural prior to geometric fidelity, we perform controlled ablation experiments on a held-out set of 500 high-quality building assets manually modeled in the CityCraft~\cite{deng2024citycraft}. Evaluation metrics include CD and F-score, computed between the generated and ground-truth 3D meshes.

\noindent \textbf{Quantitative Results.}  
Tab.~\ref{tab:geo_quantitative} summarizes the results. Sat2RealCity achieves the lowest CD and highest F-score, surpassing TRELLIS-MV even though the latter uses richer multi-view inputs. This demonstrates that explicit structural constraints, rather than simply multi-view information, are essential for geometry-faithful urban synthesis.  
Importantly, Sat2RealCity-T, which takes the satellite top-view as input, still significantly outperforms other methods. This confirms that the improvement stems from our OSM-based structural prior itself, which effectively resolves geometric ambiguities in single-view satellite imagery.

\noindent \textbf{Qualitative Results.}  
As shown in Fig.~\ref{fig:geo}, relying solely on visual cues without structural guidance often leads to noticeable distortions in roof geometry and height proportions. In contrast, our method reconstructs sharp, dimensionally consistent structures across diverse building types. These qualitative results further support the quantitative findings, highlighting that the OSM-based prior serves as a strong geometric anchor for 3D building entity generation.
\begin{figure}[t]
    \centering
    \begin{minipage}[c]{0.42\textwidth}
        \vspace{0pt} 
        \centering
        \captionof{table}{Quantitative comparison of stylistic consistency.}
        \label{tab:consis_quantitative}
        \begin{tabular}{lc}
        \toprule
        Method  & $S_{\text{regional}}$ $\uparrow$  \\
        \midrule
        Hunyuan3D-2.0-MV &  0.1236 \\
        TRELLIS-MV-M & 0.3086 \\
        TRELLIS-MV-S & 0.3027 \\
        Sat2RealCity (Ours) & \textbf{0.7885} \\
        \bottomrule
        \end{tabular}
    \end{minipage}
    \hfill
    \begin{minipage}[c]{0.54\textwidth}
        \vspace{0pt} 
        \centering
        \includegraphics[width=\linewidth]{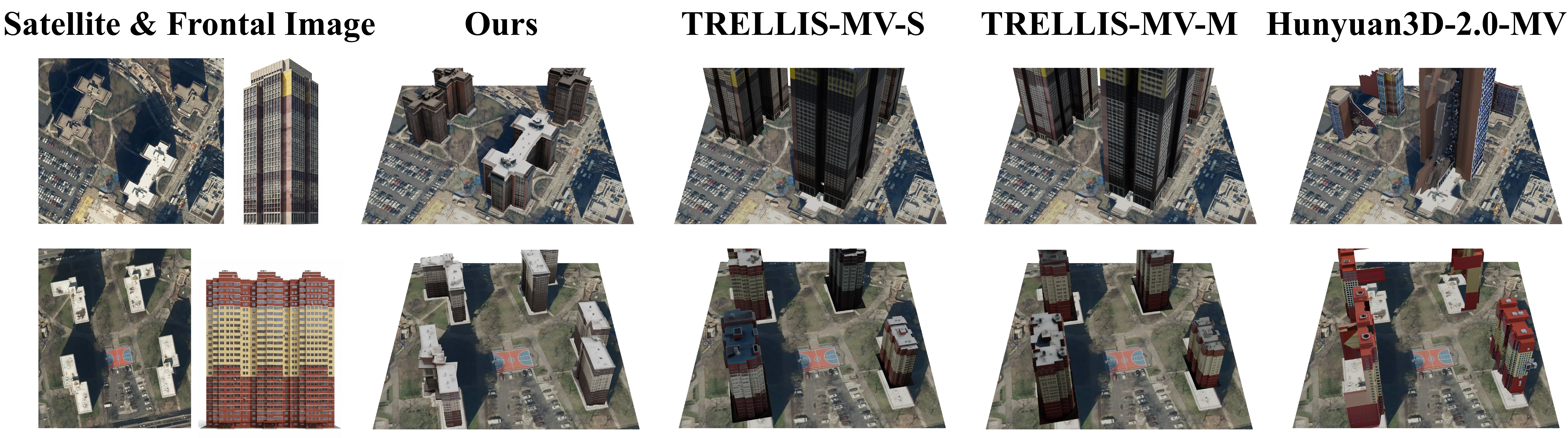} 
        \captionof{figure}{Qualitative visualization of stylistic consistency.} 
        \label{fig:app}
    \end{minipage}
\end{figure}
\subsection{Ablation Study on Stylistic Consistency}
\label{subsec:exp_consistency}
\noindent \textbf{Settings.} 
This experiment validates our \emph{appearance-guided control mechanism}, ensuring stylistic coherence across a region even when individual building geometries differ. From 111 evaluation regions, we selected 20 representative subsets with regional stylistic consistency, where buildings share a unified architectural style but differ in footprint and layout. For each region, a frontal image from our MLLM pipeline served as the \emph{shared style prompt}, and each building was generated from its \emph{unique} satellite top-view input combined with this common appearance condition.

Crucially, conventional scene-level urban generation baselines are  incompatible with this specific evaluation protocol. Because they typically represent building clusters as fused volumetric fields or inseparable global meshes, individual building entities cannot be seamlessly isolated and independently rendered. Therefore, our baselines for this entity-level consistency task comprise state-of-the-art multi-view generative models adapted for this setting: TRELLIS-MV and Hunyuan3D-2.0-MV. These models often overfit to one modality and losing either geometric alignment or stylistic coherence. To capture this trade-off, we design a composite evaluation framework with two entity-level metrics:
\begin{itemize}
\item Structural Alignment (${IoU}_{{top}}$): 
Measures how well generated buildings adhere to satellite footprints. We render each \emph{isolated} 3D building mesh top-down against a blank background, binarize it, and compute the Intersection-over-Union (IoU) with its ground-truth footprint, averaged over all buildings.

\item Appearance Consistency ($\mathit{CLIP}_{{pairwise}}$):
Quantifies the stylistic coherence \emph{among} buildings within the same region. All $N$ extracted buildings are rendered \emph{individually} at 0° elevation with a clean background. The average CLIP (ViT-L/14) cosine similarity is then computed over all unique $\frac{N(N-1)}{2}$ building pairs to strictly evaluate inter-building style uniformity.
\end{itemize}
Finally, the Regional Consistency Score ($S_{regional}$) is defined as:
\begin{equation}
S_{{regional}} = {IoU}_{{top}} \times \mathit{CLIP}_{{pairwise}},
\end{equation}

\noindent \textbf{Quantitative Results.} 
In Tab.~\ref{tab:consis_quantitative}, our method achieves the highest \(S_{\text{regional}}\), substantially surpassing all baselines. This demonstrates its superior capability in maintaining coherent regional styles while preserving geometric plausibility.

\noindent \textbf{Qualitative Results.} 
As shown in Fig.~\ref{fig:app}, our generated city regions exhibit coherent architectural styles across individual buildings with diverse shapes and layouts. Facade materials, color tones, and roof structures remain stylistically consistent while preserving accurate spatial footprints. Baselines often display inconsistent roof orientations or abrupt texture changes across adjacent buildings. These observations confirm that Sat2RealCity achieves region-level stylistic coherence while maintaining high geometric fidelity.

\begin{figure}[t]
    \centering
    \begin{minipage}[c]{0.40\textwidth}
        \vspace{0pt} 
        \centering
        \captionof{table}{Quantitative comparison with city generation methods.}
        \label{tab:citydreamer_seed3d_quantitative}
        \begin{tabular}{lcc}
        \toprule
        Method &  CLIP $\uparrow$  & DINOv2 sim$\uparrow$ \\
        \midrule
        CityDreamer  &  0.8572 & 0.6810 \\
        Seed3D &  0.8413  & 0.7344 \\
        Ours & \textbf{0.8767} & \textbf{0.7631} \\
        \bottomrule
        \end{tabular}
    \end{minipage}
    \hfill
    \begin{minipage}[c]{0.56\textwidth}
        \vspace{0pt} 
        \centering
        \includegraphics[width=\linewidth]{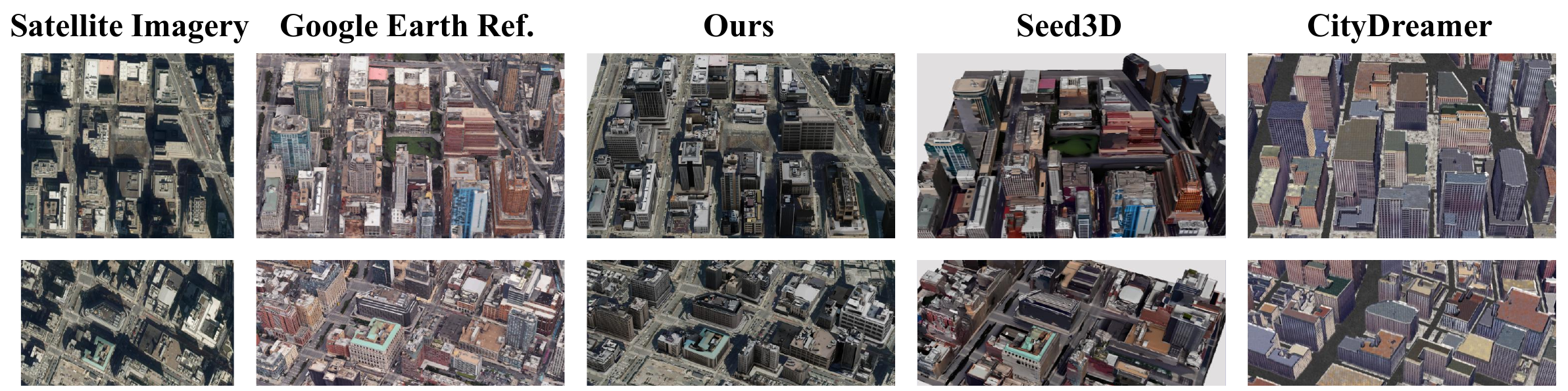} 
        \captionof{figure}{Visual comparison with city generation methods.}
        \label{fig:exp4}
    \end{minipage}
\end{figure}
\subsection{Comparison with Scene-Level City Generation Methods}
\label{subsec:exp_citygen_compare}
\noindent \textbf{Settings.} 
We further compare our framework with end-to-end city generation methods. Since recent works like Sat2City~\cite{hua2025sat2city} and EarthCrafter~\cite{liu2025earthcrafter} have not released their code, we benchmark against CityDreamer~\cite{xie2024citydreamer} and the commercial model Seed3D~\cite{seedseed3d} on a $1~\text{km}^{2}$ urban region. Crucially, CityDreamer only outputs rendered RGB frames without explicit 3D geometry. Since Seed3D cannot ingest top-down satellite inputs, we instead provide it with corresponding oblique images rendered via Google Earth Studio.

\noindent \textbf{Quantitative Results.}  
Since CityDreamer lacks geometry output, direct 3D evaluation is infeasible. Instead, we utilize 2D proxy metrics: while the CLIP score assesses semantic and stylistic appearance, DINOv2 similarity serves as an effective proxy for fine-grained geometric and structural fidelity~\cite{zhan2024general,oquab2023dinov2}. As shown in Tab.~\ref{tab:citydreamer_seed3d_quantitative}, Sat2RealCity achieves the highest scores in both, demonstrating superior performance in appearance rendering and structural preservation.

\noindent \textbf{Qualitative Results.}
Fig.~\ref{fig:exp4} presents visual comparisons. 
When conditioned on satellite imagery, our method preserves semantic information consistent with the input. Compared with reference views reconstructed from Google Earth Studio oblique images, our results exhibit superior geometric structure. While Seed3D and CityDreamer produce city layouts that roughly follow the urban structure, their outputs lack fine-grained architectural detail and physical fidelity.

\subsection{Practical Scalability Analysis}
\label{subsec:scalability}

We analyze the practical scalability of Sat2RealCity compared with scene-level generation baselines. \textbf{Memory and Parallelism.} Our framework maintains constant spatial memory complexity through iterative entity-level generation. Although auxiliary 2D models (MLLM/FLUX) temporarily increase peak VRAM, these components are easily offloadable to cloud APIs, keeping the core 3D module at only 15.7 GB. Furthermore, entity-level generation naturally supports distributed parallelism across GPUs. \textbf{Practical Rendering.} CityDreamer's volumetric rendering necessitates heavy on-the-fly computation, preventing real-time frame rates. Conversely, explicitly generated 3D polygon meshes (by ours and Seed3D) enable zero-marginal-cost real-time rendering in standard graphics engines. Thus, as scene roaming scales, our total time cost quickly undercuts implicit methods, making our framework highly feasible for massive digital twins.

\begin{table}[t]
    \centering
    \scriptsize
    \caption{Efficiency analysis on a $1 \text{km}^2$ urban region. Although our method requires offline entity generation, the resulting explicit 3D assets naturally support efficient downstream rendering. $^*$11.1 min with 8 GPUs. $^\dagger$Core 3D module VRAM.}
    \label{tab:efficiency}
    \begin{tabular}{l c c c c}
        \toprule
        Method & VRAM (GB) & Offline Gen. & Render Cost & Total (500 Frames) \\
        \midrule
        CityDreamer  & 31.1 & - & 10.9 s/frame & 91.1 min \\
        Seed3D       & - & 16.0 min & Negligible & 16.0 min \\
        \midrule
        Ours         & \textbf{15.7}$^\dagger$ / 62.6 & 88.8 min (11.1$^*$) & \textbf{Negligible} & \textbf{88.8 min} \\
        \bottomrule
    \end{tabular}
    \vspace{-8pt}
\end{table}

\section{Conclusion}
\label{sec:conclusion}
We present Sat2RealCity, a building-entity-level framework for grounded 3D urban generation from satellite imagery. Leveraging object-level 3D generative priors, it enables compositional urban synthesis while preserving map-derived geographic structures and regional stylistic consistency. Supported by the BuildVerse3D dataset, it integrates OSM-based spatial grounding, appearance-guided controllable generation, and an MLLM-powered semantic pipeline. Extensive experiments demonstrate effective geographic alignment, stylistic coherence, and plausible urban asset synthesis compared to existing urban and 3D asset generation approaches. This work provides a practical step toward scalable geographically grounded urban generation and digital twin content creation.

\bibliographystyle{splncs04}
\bibliography{main}

\end{document}